\documentclass{article}
\usepackage{spconf,amsmath,graphicx,hyperref}

\usepackage{booktabs,multirow,amsfonts}

\usepackage{enumitem}

\title{FOCA: Frequency-Oriented Cross-Domain Forgery Detection, Localization and Explanation via Multi-Modal Large Language Model}

\name{Zhou Liu$^{1,2}$, Tonghua Su$^{1,3,4}$, Hongshi Zhang$^1$, Fuxiang Yang$^1$, Donglin Di$^{1,2}$, Yang Song$^5$, Lei Fan$^{2,5\star}$}
\address{$^1$Harbin Institute of Technology, Harbin, China\\$^2$DZ-Matrix\\
$^3$Guangdong Laboratory of Artificial Intelligence and Digital Economy (SZ), Shenzhen, China\\
$^4$Chongqing Research Institute of HIT, Chongqing, China\\
$^5$University of New South Wales, Sydney, Australia}

\begin{document}
\maketitle
\begin{abstract}

Advances in image tampering techniques, particularly generative models, pose significant challenges to media verification, digital forensics, and public trust. Existing image forgery detection and localization (IFDL) methods suffer from two key limitations: over-reliance on semantic content while neglecting textural cues, and limited interpretability of subtle low-level tampering traces. To address these issues, we propose FOCA, a multimodal large language model–based framework that integrates discriminative features from both the RGB spatial and frequency domains via a cross-attention fusion module. This design enables accurate forgery detection and localization while providing explicit, human-interpretable cross-domain explanations. We further introduce FSE-Set, a large-scale dataset with diverse authentic and tampered images, pixel-level masks, and dual-domain annotations. Extensive experiments show that FOCA outperforms state-of-the-art methods in detection performance and interpretability across both spatial and frequency domains.
\end{abstract}
\begin{keywords}
Image Forgery Detection, Multimodal Large Language Model, Frequency Domain Analysis
\end{keywords}
\section{Introduction}

Image forgery detection and localization (IFDL) aims to determine whether an image has been tampered with and to identify manipulated regions. It plays a critical role in multimedia forensics, digital evidence authentication, and combating misinformation. With the rapid advancement of generative models~\cite{rombach2022high}, highly realistic synthetic images and seamless edits can now closely mimic natural image statistics, posing unprecedented challenges to traditional IFDL methods~\cite{PSCC-Net} and threatening media authenticity.

Existing IFDL approaches can be broadly categorized into two paradigms. The first relies on pretrained image encoders, such as MVSS-Net~\cite{MVSS_2022TPAMI} and HiFi-Net~\cite{Hifi-Net}, which often incorporate auxiliary cues from noise or frequency domains. These methods are computationally efficient but typically produce only detection scores or tampering masks, offering limited interpretability. The second paradigm consists of multimodal large language models (MLLMs)~\cite{SIDA,zang2025sage,sun2025llapa}, which leverage large-scale semantic knowledge to identify high-level inconsistencies. Representative works include SIDA~\cite{SIDA}, which constructs a cross-modal training dataset, and ForgeryGPT~\cite{liu2024forgerygpt}, which augments LLMs with mask-aware forgery extraction. Despite their strong semantic reasoning capabilities, existing MLLM-based methods operate exclusively in the RGB domain and often overlook subtle forensic traces, such as texture irregularities and high-frequency artifacts.

In this paper, we aim to equip MLLMs with frequency-domain awareness for a more interpretable IFDL framework. To this end, we propose a Frequency-Oriented Cross-domain forgery detection, termed FOCA. It integrates RGB- and frequency-domain feature representations through a Frequency Attention Fusion (FAF) module. FAF leverages cross-attention and contrastive learning to dynamically fuse spatial and wavelet-frequency features, enabling the model to associate semantic inconsistencies with forensic traces and thereby improving both detection and localization performance. Furthermore, we construct a large-scale dataset named FSE-Set to fine-tune MLLMs. It contains diverse manipulated and authentic images with pixel-level masks and provides both RGB- and frequency-domain explanations to facilitate interpretable evaluation.

Our methodological contributions are as follows:
We propose FOCA, the first MLLM-based framework that integrates semantic reasoning with frequency-domain forensic cues for interpretable image forgery detection and localization.
We construct the FSE-Set dataset, including multi-domain annotations that enable explainable forgery analysis across both spatial and frequency domains.
Extensive experiments demonstrate that FOCA outperforms state-of-the-art methods in detection accuracy while providing human-interpretable explanations in both RGB and frequency domains.

\begin{figure*}[t]
    \includegraphics[width=0.98\textwidth]{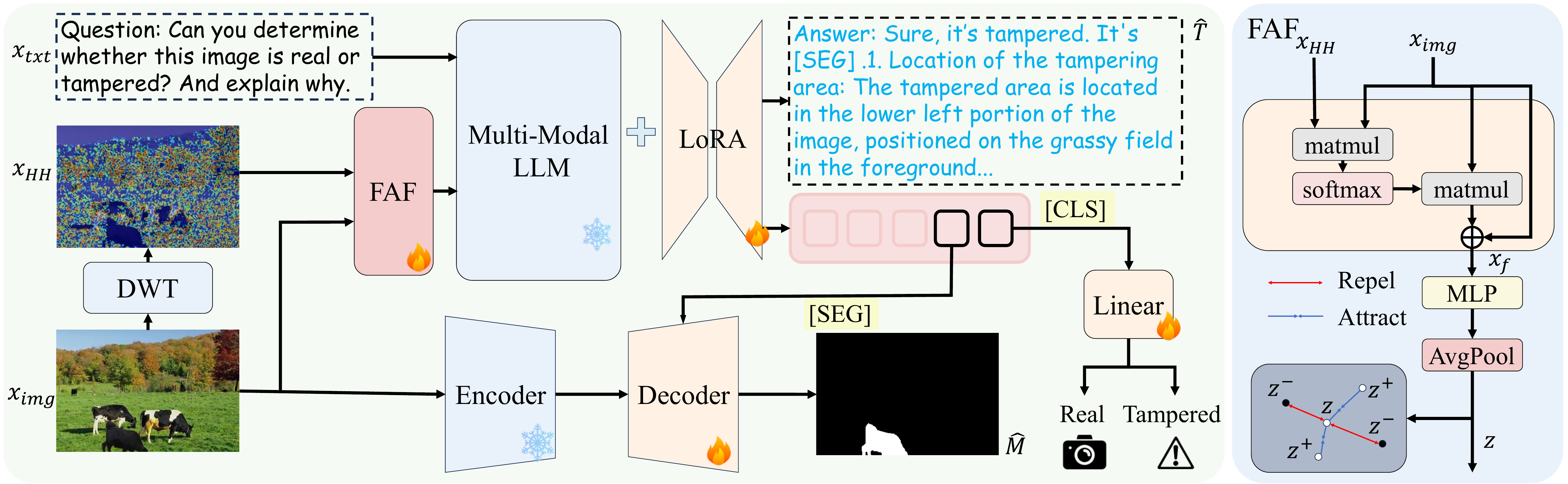}
    \centering
    \caption{Overview of FOCA. It receives an image $x_{img}$ and a text instruction $x_{txt}$ and produces a textual response. FOCA uses the [CLS] token for tamper classification and the [SEG] token for pixel-level mask generation.
    }
    \label{fig:architecture}
\end{figure*}

\section{Method}

\subsection{Overview}

Given a text instruction $x_{txt}$ and an input image $x_{img} \in \mathbb{R}^{H \times W \times 3}$, our goal is to jointly predict three outputs: the detection result $\hat{D}$, the tampered area mask $\hat{M}$ and a textual explanation of tampering artifacts $\hat{T}$. The overall architecture is shown in Fig.~\ref{fig:architecture}. Our FOCA comprises three components:
\begin{itemize}[itemsep=0pt, topsep=0pt, leftmargin=1.5em]
    \item \textit{Frequency Attention Fusion (FAF)}. To enhance sensitivity to tampering artifacts, we extract high-frequency components from $x_{img}$ and fuse them with its original image, yielding a fused feature $x_f$.
    \item \textit{MLLM backbone $\mathcal{F}$}. The fused features $x_f$, together with the text instruction $x_{txt}$, are passed into the MLLM $\mathcal{F}$ to generate the textual explanation $\hat{T}$. To enable detection and localization~\cite{Lisa}, we extend the vocabulary of the MLLM with two special tokens, [SEG] and [CLS], corresponding to segmentation and detection tasks.
    \item \textit{Segmentation module}. We extract the last-layer embeddings $\widetilde{h}_{seg}$ and $\widetilde{h}_{cls}$ of [SEG] and [CLS], respectively. The $\widetilde{h}_{seg}$ is projected via an MLP layer $\gamma$ to produce $h_{seg}$, while $\widetilde{h}_{cls}$ is passed through a classifier $\phi$ to yield the detection result $\hat{D}$:
    \begin{equation}
     \hat{h}_{seg}=\gamma(\widetilde{h}_{seg}), \quad
     \hat{D}=\phi(\widetilde{h}_{cls}).
     \label{eq:result_D}
     \end{equation}
\end{itemize}   
Meanwhile, a frozen image encoder $S_{enc}$ extracts visual features $f$ from $x_{img}$, which are combined with $h_{seg}$ and decoded by $S_{dec}$ to generate the pixel-level mask $\hat{M}$.

During training, the image encoder and MLLM parameters are frozen. We train only the FAF module and the segmentation decoder. The MLLM is fine-tuned with LoRA~\cite{hu2022lora}, which updates a small number of low-rank matrices, ensuring efficiency while retaining the pretrained model capacity.

\subsection{Frequency Attention Fusion}

Prior studies~\cite{li2025improving,yi2022approximate} have shown that tampering often leaves subtle traces in frequency components. We adopt the Discrete Wavelet Transform (DWT) for high-frequency feature extraction, since DWT not only isolates frequency information but also preserves spatial structures.

Specifically, DWT decomposes the input image $x_{img}$ into four sub-bands $x_{LL}, x_{LH}, x_{HL}, x_{HH} \in \mathbb{R}^{H/2 \times W/2 \times 3}$, where ``L''and ``H'' denote low-/high-pass filters capturing smooth regions and edges, respectively. Among them, the $x_{HH}$ sub-band demonstrates strong generalization capability and effectively reveals subtle tampering artifacts~\cite{li2025improving}. This sub-band is integrated with spatial features through the FAF module. The FAF employs a cross-attention mechanism in which $x_{HH}$ acts as the query, while the original image $x_{img}$ serves as the key and value. The process is defined as:
\begin{equation}
A=\operatorname{softmax}\left(\frac{(\mathbf{w}_qx_{HH})(\mathbf{w}_kx_{img})^\top}{\sqrt{d_k}}\right)\mathbf{w}_vx_{img},
\label{eq:attn}
\end{equation}
where $A$ is the attention score matrix, $\mathbf{w}_q$, $\mathbf{w}_k$, and $\mathbf{w}_v$ are trainable projection matrices, and $d_k$ is the dimensionality of the key vectors. This design adaptively retrieves structurally relevant regions from spatial features under the guidance of high-frequency cues. The attended features are then fused with spatial features via a residual connection, thereby retaining low- and mid-frequency information while emphasizing tampering-sensitive details.

To stabilize training and enhance robustness, we introduce a residual connection defined as:
\begin{equation}
x_f=\operatorname{proj}(A)+x_{img},
\label{eq:residual}
\end{equation}
where $\operatorname{proj}(\cdot)$ denotes a linear projection. It selectively amplifies high-frequency details that are critical for tampering localization, while suppressing noise propagation.

Inspired by recent contrastive learning strategies~\cite{fan2025salvaging}, we introduce an auxiliary contrastive objective to enhance discriminative feature learning. Specifically, for each input image, the fused features ${x_f}$ are first passed through spatial average pooling and then projected into a latent space via a two-layer MLP with ReLU activation. The resulting embedding is normalized to yield a $d$-dimensional vector:
\begin{equation}
z=\operatorname{Norm}(\operatorname{MLP}(\operatorname{AvgPool}({x_f}))).
\label{eq:z}
\end{equation}
Given a batch of $N$ images, each sample forms a positive pair with itself, while all other samples serve as negatives. We compute the scaled cosine similarity between embeddings as:
\begin{equation}
s_{ij} = \frac{z_i \cdot z_j}{\tau \|z_i\| \|z_j\|},
\label{eq:similarity}
\end{equation}
where $\tau$ is a temperature controlling the distribution sharpness. The objective follows the InfoNCE formulation:
\begin{equation}
\mathcal{L}_{cl} = -\frac{1}{N} \sum_{i=1}^{N} \log \frac{\exp(s_{ii})}{\sum_{j=1}^{N} \exp(s_{ij})}.
\label{eq:contrastive_loss}
\end{equation}
This loss maximizes agreement between positive pairs while minimizing similarity with negatives, thereby enforcing discriminative, tampering-aware representations that complement the primary detection and localization objectives.

\subsection{FSE-Set Dataset}

Existing IFDL datasets~\cite{Casiav1,Columbia,SIDA} provide tampering annotations but lack textual explanations and coverage of advanced manipulation techniques. To bridge these gaps, following prior dataset construction methodologies~\cite{SIDA,zang2025sage,sun2025llapa}, we introduce FSE-Set, a new dataset designed to support cross-domain image forgery analysis in both spatial and frequency domains.

\begin{figure}[t]
    \includegraphics[width=\linewidth]{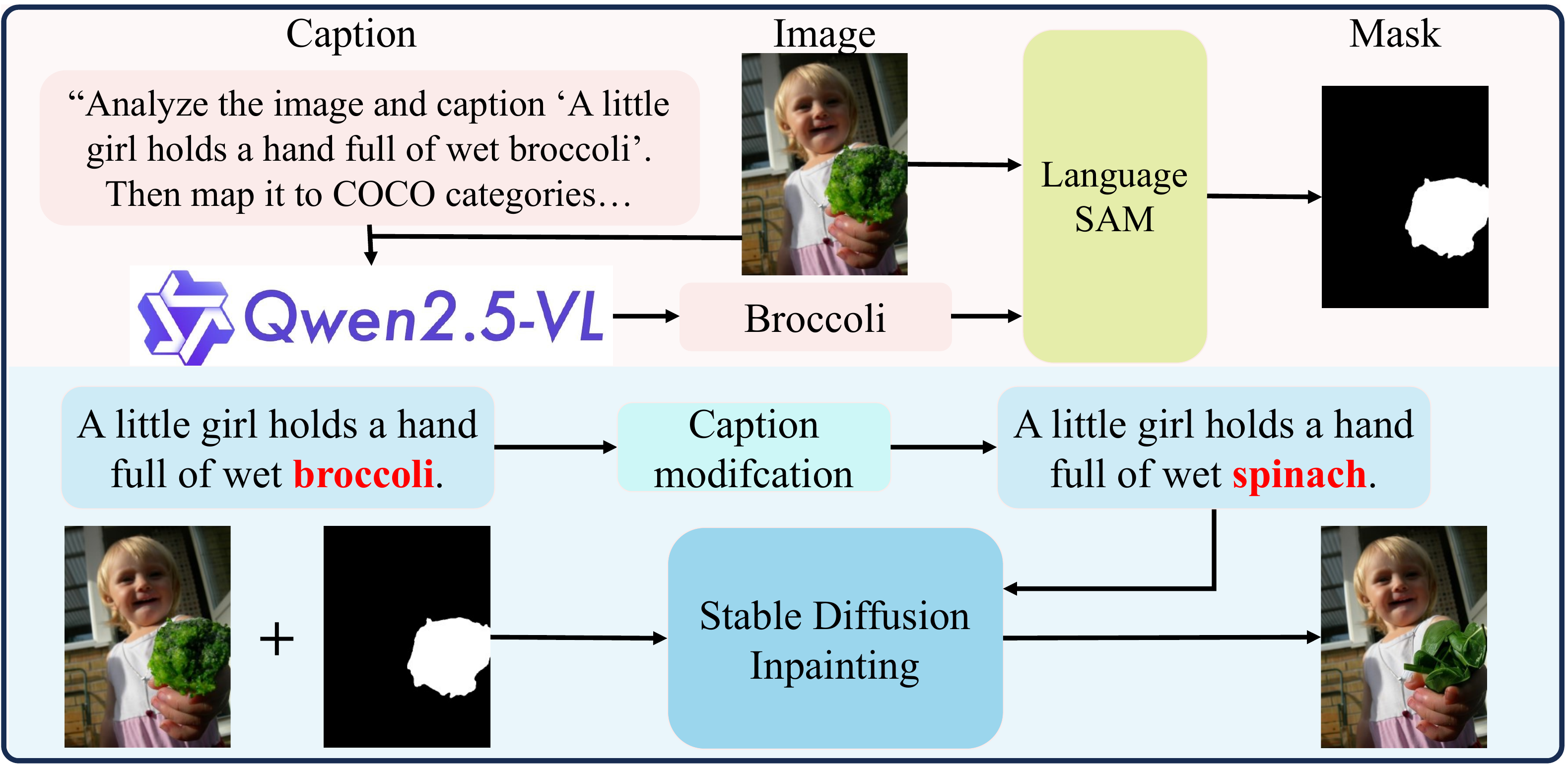}
    \caption{AI-Generated manipulations pipeline.}
    \label{fig:generation}
\end{figure}

FSE-Set consists of 50K real images from ImageNet~\cite{imagenet} and 50K tampered images derived from COCO~\cite{coco}, including 25K traditional manipulations (splicing and copy-move) and 25K AI-generated edits. As shown in Fig.~\ref{fig:generation}, tampered samples are constructed by object extraction with Qwen2.5-VL~\cite{Qwen2.5-VL}, mask generation using Language-SAM~\cite{lang-segment-anything}, and image modification via Stable Diffusion inpainting~\cite{rombach2022high}.

To provide dual-domain insights for tampering analysis, we further employ Claude~\cite{claude} to jointly analyze the RGB image and its HH frequency sub-band, producing fine-grained explanations in both domains for training and evaluation.

\subsection{Training Objectives}

Our framework jointly optimizes two complementary loss functions: the prediction loss and the contrastive loss. The overall loss is defined as:
\begin{equation} 
\mathcal{L} = \mathcal{L}_{pred} + \lambda_{c}\mathcal{L}_{cl},
\label{eq:final_loss}
\end{equation}
where $\lambda_{c}$ balances the prediction and contrastive components. The prediction loss $\mathcal{L}_{pred}$ consists of three terms:
\begin{equation} 
\mathcal{L}_{pred} = \mathcal{L}_{{t}} + \mathcal{L}_{cls} 
+ \mathcal{L}_{mask} ,
\label{eq:pred_loss}
\end{equation}
where the text generation loss $\mathcal{L}_{t}$ and classification loss $\mathcal{L}_{cls}$ are both formulated as cross-entropy losses, measuring discrepancies between the generated text $\hat{T}$ and ground truth captions $T$, as well as between the predicted category $\hat{D}$ and true label $D$.

For segmentation, we adopt a composite objective combining binary cross-entropy (BCE) and Dice loss to optimize the predicted mask $\hat{M}$ against the ground truth mask $M$:
\begin{equation}
\mathcal{L}_{mask} = \lambda_{bce}\mathcal{L}_{BCE}(\hat{M}, M) + \mathcal{L}_{dice}(\hat{M}, M),
\label{eq:mask_loss}
\end{equation}
where $\lambda_{bce}$ and $\lambda_{dice}$ are balancing coefficients. This formulation encourages pixel-level accuracy while ensuring structural consistency of predicted tampered regions.

\section{Experiments}
\subsection{Experimental built upon}

We conducted experiments on FSE-Set, CASIAv1~\cite{Casiav1}, and Columbia~\cite{Columbia}. Our model was based on LISA-7B~\cite{Lisa} as the MLLM backbone and SAM~\cite{sam} as the vision backbone. The model was fine-tuned using LoRA~\cite{hu2022lora} and a dropout rate of 0.05. Training was conducted on 4× NVIDIA H200 GPUs. The loss weights were set to $\lambda_{c}=0.5$ and $\lambda_{bce}=2.0$.

\textbf{Evaluation Metrics.} For detection, we report image-level accuracy and F1 score. For localization, we adopt F1 and IoU. To evaluate artifact explanation quality, we use ROUGE-L and cosine similarity (CSS). Additionally, inspired by the LLM-as-a-Judge paradigm~\cite{llm_judge}, we employ GPT-4o~\cite{gpt4o} as an evaluator. A carefully designed prompt guides GPT-4o to impartially compare the model’s explanation against references, analyze similarities, differences, and errors, before assigning a quantitative score from 1 to 10.

\subsection{Results and Discussion}
\textbf{Detection Evaluation.} To evaluate the detection capability of our model, we conducted comparative experiments on FSE-Set with four state-of-the-art traditional detection methods that are retrained on our FSE-Set, namely CnnSpott~\cite{CNNDetection}, Fusing~\cite{ju2022fusing}, Uni-vFD~\cite{UnivFD}, and DRCT~\cite{drct2024}. The detailed comparison results are presented in Table \ref{tab:detection_comparison}. Our method achieved the best performance across most metrics, especially in F1 scores for both Real (96.2\%) and Tampered (96.3\%) classes, and overall accuracy (96.2\%). While Fusing~\cite{ju2022fusing} attained slightly higher results in Real accuracy (98.1\%), our approach demonstrates a stronger balance between the two classes and superior performance in tampered-image detection, highlighting its robustness in practical forgery scenarios.

\begin{table}[t]
  \centering
  \caption{Comparison with deepfake detection methods.}
  \label{tab:detection_comparison}
  \begin{tabular}{ccccccc}
    \toprule
    \multirow{2}{*}{Methods} & \multicolumn{2}{c}{Real} & \multicolumn{2}{c}{Tampered} & \multicolumn{2}{c}{Overall} \\
    \cmidrule(lr){2-3} \cmidrule(lr){4-5} \cmidrule(lr){6-7}
     & Acc & F1 & Acc & F1 & Acc & F1\\
    \midrule
    CnnSpott~\cite{CNNDetection} & 93.9 & 92.1 & 90.4 & 92.1 & 92.1 & 92.1\\
    Fusing~\cite{ju2022fusing}   & \textbf{98.1} & 94.7 & 91.4 & 94.6 & 94.6 & 94.6\\
    UnivFD~\cite{UnivFD}   & 88.7 & 88.0 & 87.9 & 88.6 & 88.6 & 88.3\\
    DRCT~\cite{drct2024}     & 96.7 & 93.2 & 89.5 & 92.9 & 93.0 & 93.0\\
    \hline
    Our      & 97.1 & \textbf{96.2} & \textbf{95.4} & \textbf{96.3} & \textbf{96.2} & \textbf{96.2}\\
    \bottomrule
  \end{tabular}
\end{table}

\begin{table}[t]
  \centering
  \caption{Comparison of MLLM-based methods in detection performance and tamper localization performance.}
  \label{tab:combined_comparison}
  \resizebox{\linewidth}{!}{%
  \begin{tabular}{ccccccccc}
    \toprule
     \multirow{2}{*}{Method} & \multicolumn{2}{c}{Detection} & \multicolumn{2}{c}{FSE-Set} & \multicolumn{2}{c}{CASIA v1} & \multicolumn{2}{c}{Columbia} \\
    \cmidrule(lr){2-3} \cmidrule(lr){4-5} \cmidrule(lr){6-7} \cmidrule(lr){8-9}
    & Acc & F1 & IoU & F1 & IoU & F1 & IoU & F1 \\
    \midrule
    LISA~\cite{Lisa}     & -- & -- & 32.5 & 49.0 & 6.1 & 11.6 & 13.7 & 24.1 \\
    DS~\cite{wu2024deepseekvl2} & 49.2 & 34.5 & 9.5 & 17.3 & 7.9 & 14.7 & 25.6 & 40.1 \\
    Qwen~\cite{Qwen2.5-VL} & 57.3 & 63.9 & 23.4 & 37.9 & 15.0 & 26.0 & 45.4 & 62.4 \\
    IVL3~\cite{internvl3} & 68.4 & 79.5 & 7.6 & 14.0 & 6.6 & 12.4 & 30.4 & 46.6 \\
    SIDA~\cite{SIDA}  & 95.6 & 95.6 & 47.9 & 64.7 & \textbf{37.6} & \textbf{54.7} & 60.2 & 74.4 \\
    \hline
    Ours     & \textbf{96.2} & \textbf{96.2} & \textbf{48.6} & \textbf{65.4} & 34.4 & 51.2 & \textbf{60.8} & \textbf{75.6} \\
    \bottomrule
  \end{tabular}%
  }
\end{table}

We further compared our method with four MLLM-based approaches: InternVL3 (IVL3)~\cite{internvl3}, Qwen2.5-VL (Qwen)~\cite{Qwen2.5-VL}, DeepSeekVL2 (DS), and SIDA~\cite{SIDA}. For InternVL3, Qwen2.5-VL, and DeepSeekVL2, we used their released pre-trained models, while SIDA was retrained on our FSE-Set. The results are shown in Table \ref{tab:combined_comparison}. Our model achieved the best performance in detection, with both Accuracy and F1 score reaching 96.2\%, outperforming all other methods. The closest competitor, SIDA, achieved 95.6\%, while other models fell further behind. The key advantage of our approach lies in feeding both RGB and frequency-domain representations into the MLLM. By leveraging high-frequency information, our framework captures subtle forensic cues that are often invisible in the spatial domain alone.

\begin{figure}[t]
    \includegraphics[width=\linewidth]{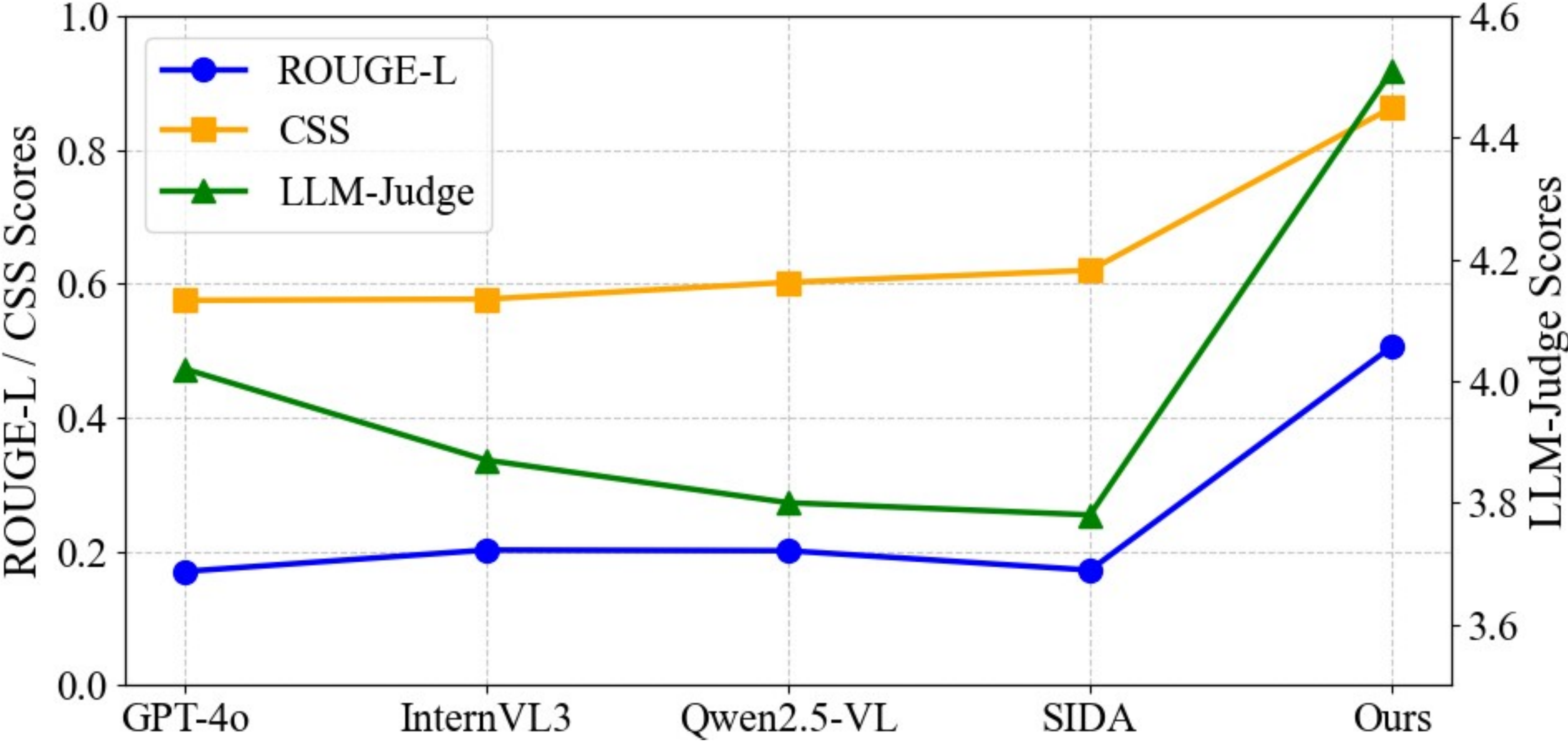}
    \caption{Comparative results of the pre-trained MLLMs and FOCA in tampering explanation capabilities on the FSE-Set.}
    \label{fig:plot}
\end{figure}

\textbf{Localization Evaluation.}

The localization evaluation experiment assesses FOCA’s ability to precisely identify pixel-level tampered regions on the FSE-Set, CASIA v1, and Columbia datasets. As shown in Table~\ref{tab:combined_comparison}, FOCA achieves SOTA performance on FSE-Set (+0.7 IoU and +0.7 F1 over SIDA) and Columbia (+0.6 IoU and +1.2 F1 over SIDA). On CASIA v1, it slightly trails SIDA in IoU and F1 but remains highly competitive. All other MLLM-based methods perform significantly worse than FOCA and SIDA across all datasets. FOCA’s superior localization performance is driven by the FAF module, which dynamically associates spatial semantic inconsistencies with high-frequency forensic traces to improve region-level precision. Moreover, FOCA provides interpretable explanations from both spatial and frequency domains, enhancing the credibility of its localization results.

\textbf{Explanation Evaluation.} To evaluate the model's interpretability and its ability to generate natural language explanations, we compared it with GPT-4o, InternVL3, Qwen2.5-VL, and SIDA on the FSE-Set dataset. We used ROUGE-L for surface-level structural alignment and CSS for semantic equivalence to jointly assess lexical coherence and contextual fidelity. Additionally, we employed the LLM-Judge score, which provides a human-like qualitative assessment of explanation quality on a 10-point scale. As shown in Fig.~\ref{fig:plot}, our model achieved top performance in ROUGE-L, CSS, and LLM-Judge, demonstrating effectiveness in generating high-quality tampering explanations.

\section{Conclusion}
This paper introduces frequency-domain interpretation to the IFDL task and presents a dataset of 100,000 images with spatial and frequency analysis. We propose FOCA, a multimodal framework that detects, localizes, and explains tampering artifacts in both domains, advancing image forensics and supporting efforts against misinformation.

\section{Acknowledgents}
This work was supported by the National Natural Science Foundation of China (Grant No. 62277011),  National Key Research and Development Program of China (Grant No. GG-2024-01-02), Open Research Fund from Guangdong Laboratory of Artificial Intelligence and Digital Economy (SZ) (Grant No.GML-KF-24-18) and Project of Chongqing MEITC(Grant No. YJX-2025001001009).

\newpage

\bibliographystyle{IEEEbib}
\bibliography{strings,refs}

\begin{thebibliography}{10}

\bibitem{rombach2022high}
Robin Rombach, Andreas Blattmann, et~al.,
\newblock ``High-resolution image synthesis with latent diffusion models,''
\newblock in {\em CVPR}, 2022, pp. 10684--10695.

\bibitem{PSCC-Net}
Xiaohong Liu, Yaojie Liu, et~al.,
\newblock ``Pscc-net: Progressive spatio-channel correlation network for image manipulation detection and localization,''
\newblock {\em TCSVT}, 2022.

\bibitem{MVSS_2022TPAMI}
Chengbo Dong, Xinru Chen, et~al.,
\newblock ``Mvss-net: Multi-view multi-scale supervised networks for image manipulation detection,''
\newblock {\em TPAMI}, pp. 1--14, 2022.

\bibitem{Hifi-Net}
Xiao Guo, Xiaohong Liu, et~al.,
\newblock ``Hierarchical fine-grained image forgery detection and localization,''
\newblock in {\em CVPR}, 2023, pp. 3155--3165.

\bibitem{SIDA}
Zhenglin Huang, Jinwei Hu, et~al.,
\newblock ``Sida: Social media image deepfake detection, localization and explanation with large multimodal model,''
\newblock in {\em CVPR}, 2025, pp. 28831--28841.

\bibitem{zang2025sage}
Guoxin Zang, Xue Li, et~al.,
\newblock ``Sage: A visual language model for anomaly detection via fact enhancement and entropy-aware alignment,''
\newblock in {\em ACM MM}, 2025, pp. 5030--5039.

\bibitem{sun2025llapa}
Shibo Sun, Xue Li, et~al.,
\newblock ``Llapa: A vision-language model framework for counterfactual-aware procedural planning,''
\newblock in {\em ACM MM}, 2025, pp. 5020--5029.

\bibitem{liu2024forgerygpt}
Jiawei Liu, Fanrui Zhang, et~al.,
\newblock ``Forgerygpt: Multimodal large language model for explainable image forgery detection and localization,''
\newblock {\em arXiv:2410.10238}, 2024.

\bibitem{Lisa}
Xin Lai, Zhuotao Tian, et~al.,
\newblock ``Lisa: Reasoning segmentation via large language model,''
\newblock in {\em CVPR}, 2024, pp. 9579--9589.

\bibitem{hu2022lora}
Edward~J Hu, Yelong Shen, et~al.,
\newblock ``Lora: Low-rank adaptation of large language models.,''
\newblock {\em ICLR}, vol. 1, no. 2, pp. 3, 2022.

\bibitem{li2025improving}
Ouxiang Li, Jiayin Cai, et~al.,
\newblock ``Improving synthetic image detection towards generalization: An image transformation perspective,''
\newblock in {\em ACM SIGKDD}, 2025, pp. 2405--2414.

\bibitem{yi2022approximate}
Kai Yi, Jialin Chen, et~al.,
\newblock ``Approximate equivariance so (3) needlet convolution,''
\newblock in {\em Proc. TAGL Workshops}. PMLR, 2022, pp. 189--198.

\bibitem{fan2025salvaging}
Lei Fan, Junjie Huang, et~al.,
\newblock ``Salvaging the overlooked: Leveraging class-aware contrastive learning for multi-class anomaly detection,''
\newblock in {\em ICCV}, 2025, pp. 21419--21428.

\bibitem{Casiav1}
Jing Dong, Wei Wang, et~al.,
\newblock ``{CASIA} image tampering detection evaluation database,''
\newblock in {\em ChinaSIP}, July 2013.

\bibitem{Columbia}
Y.-F. Hsu and S.-F. Chang,
\newblock ``Detecting image splicing using geometry invariants and camera characteristics consistency,''
\newblock in {\em ICME}, 2006.

\bibitem{imagenet}
Jia Deng, Wei Dong, et~al.,
\newblock ``Imagenet: A large-scale hierarchical image database,''
\newblock in {\em CVPR}, 2009, pp. 248--255.

\bibitem{coco}
Tsung-Yi Lin, Michael Maire, et~al.,
\newblock ``Microsoft coco: Common objects in context,''
\newblock in {\em ECCV}, 2014, pp. 740--755.

\bibitem{Qwen2.5-VL}
Shuai Bai, Keqin Chen, et~al.,
\newblock ``Qwen2.5-vl technical report,''
\newblock {\em arXiv:2502.13923}, 2025.

\bibitem{lang-segment-anything}
lang-sam team,
\newblock ``lang-segment-anything,'' 2024,
\newblock \url{https://github.com/luca-medeiros/lang-segment-anything}.

\bibitem{claude}
Anthropic,
\newblock ``Claude sonnet 4,'' 2025,
\newblock \url{https://www.anthropic.com/claude/sonnet}.

\bibitem{sam}
Alexander Kirillov, Eric Mintun, et~al.,
\newblock ``Segment anything,''
\newblock in {\em ICCV}, 2023, pp. 4015--4026.

\bibitem{llm_judge}
Haitao Li, Qian Dong, et~al.,
\newblock ``Llms-as-judges: a comprehensive survey on llm-based evaluation methods,''
\newblock {\em arXiv:2412.05579}, 2024.

\bibitem{gpt4o}
Josh Achiam, Steven Adler, et~al.,
\newblock ``Gpt-4 technical report,''
\newblock {\em arXiv:2303.08774}, 2023.

\bibitem{CNNDetection}
Sheng-Yu Wang, Oliver Wang, et~al.,
\newblock ``Cnn-generated images are surprisingly easy to spot... for now,''
\newblock in {\em CVPR}, 2020, pp. 8695--8704.

\bibitem{ju2022fusing}
Yan Ju, Shan Jia, et~al.,
\newblock ``Fusing global and local features for generalized ai-synthesized image detection,''
\newblock in {\em ICIP}, 2022, pp. 3465--3469.

\bibitem{UnivFD}
Utkarsh Ojha, Yuheng Li, et~al.,
\newblock ``Towards universal fake image detectors that generalize across generative models,''
\newblock in {\em CVPR}, 2023, pp. 24480--24489.

\bibitem{drct2024}
Baoying Chen, Jishen Zeng, et~al.,
\newblock ``Drct: Diffusion reconstruction contrastive training towards universal detection of diffusion generated images,''
\newblock in {\em ICML}, 2024.

\bibitem{wu2024deepseekvl2}
Zhiyu Wu, Xiaokang Chen, et~al.,
\newblock ``Deepseek-vl2: Mixture-of-experts vision-language models for advanced multimodal understanding,'' 2024.

\bibitem{internvl3}
Jinguo Zhu, Weiyun Wang, et~al.,
\newblock ``Internvl3: Exploring advanced training and test-time recipes for open-source multimodal models,'' 2025.

\end{thebibliography}

\end{document}